\documentclass[10pt,twocolumn,letterpaper]{article}

\usepackage{tikz}
\usepackage{pgfplots}
\pgfplotsset{compat=1.18} 
\usepackage{xcolor} 
\definecolor{appleblue}{RGB}{86,180,233}
\definecolor{appleteal}{RGB}{0,158,115}
\definecolor{applemustard}{RGB}{240,228,66}
\definecolor{applepurple}{RGB}{204,121,167}
\definecolor{appleorange}{RGB}{230,159,0}
\usepackage{multirow}
\usepackage{placeins} 
\usepackage{graphicx}
\usepackage{float}
\usepackage[table]{xcolor}

\usepackage[pagenumbers]{wacv} 

\definecolor{wacvblue}{rgb}{0.21,0.49,0.74}
\usepackage[pagebackref,breaklinks,colorlinks,allcolors=wacvblue]{hyperref}

\def\wacvPaperID{2992} 
\def\confName{WACV}
\def\confYear{2026}

\definecolor{applegreen}{rgb}{0.55, 0.71, 0.0}
\definecolor{atomictangerine}{rgb}{1.0, 0.6, 0.4}

\definecolor{darkgreen}{rgb}{0.0, 0.5, 0.0}

\DeclareRobustCommand{\greenCircledNumber}[1]{%
    \tikz[baseline={(0,-.3em)}]{%
        \node[circle, fill=darkgreen, inner sep=0.8pt, 
              minimum size=1.0em, text=white, font=\footnotesize] {#1};}%
}

\title{PhyEduVideo: A Benchmark for Evaluating Text-to-Video Models for Physics Education}

\author{
Megha Mariam K.M\\
IIIT Hyderabad, India\\
{\tt\small megha.km@research.iiit.ac.in}
\and
Aditya Arun\\
Adobe MDSR, India\\
{\tt\small adityaarun@adobe.com}
\and
Zakaria Laskar\\
IISER Thiruvananthapuram, India \\
{\tt\small zakaria.laskar@iisertvm.ac.in}
\and
C.V. Jawahar\\
IIIT Hyderabad, India\\
{\tt\small jawahar@iiit.ac.in}
}

\begin{document}
\maketitle

\begin{abstract}
Generative AI models, particularly Text-to-Video (T2V) systems, offer a promising avenue for transforming science education by automating the creation of engaging and intuitive visual explanations. In this work, we take a first step toward evaluating their potential in physics education by introducing a dedicated benchmark for explanatory video generation. The benchmark is designed to assess how well T2V models can convey core physics concepts through visual illustrations. Each physics concept in our benchmark is decomposed into granular teaching points, with each point accompanied by a carefully crafted prompt intended for visual explanation of the teaching point. T2V models are evaluated on their ability to generate accurate videos in response to these prompts. Our aim is to systematically explore the feasibility of using T2V models to generate high-quality, curriculum-aligned educational content—paving the way toward scalable, accessible, and personalized learning experiences powered by AI. Our evaluation reveals that current models produce visually coherent videos with smooth motion and minimal flickering, yet their conceptual accuracy is less reliable. Performance in areas such as mechanics, fluids, and optics is encouraging, but models struggle with electromagnetism and thermodynamics, where abstract interactions are harder to depict. These findings underscore the gap between visual quality and conceptual correctness in educational video generation. We hope this benchmark helps the community close that gap and move toward T2V systems that can deliver accurate, curriculum-aligned physics content at scale. The benchmark and accompanying codebase are publicly available at \url{https://github.com/meghamariamkm/PhyEduVideo}.
\end{abstract}

\begin{figure*}[t]
    \centering
    \includegraphics[width=1\linewidth]{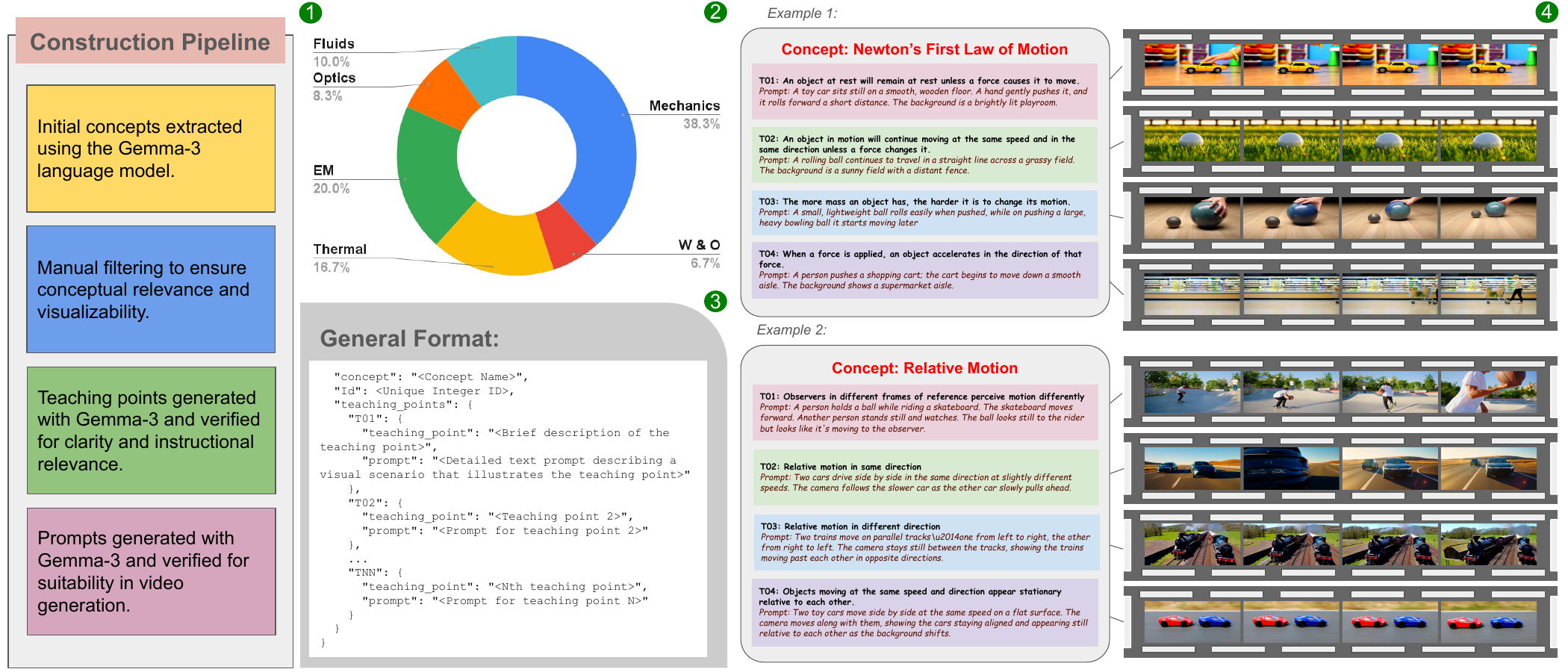}
    \caption{\textbf{Overview of the PhyEduVideo Benchmark.} 
\greenCircledNumber{1} The construction pipeline, from concept extraction to prompt generation. 
\greenCircledNumber{2} Concept distribution across five major physics domains: \textit{Mechanics}, \textit{Electromagnetism} (\textit{EM}), \textit{Optics}, \textit{Thermodynamics} (Thermo), \textit{Fluids}, and \textit{Waves} \& \textit{Oscillations} ( \textit{W}\&\textit{O}). 
\greenCircledNumber{3} Standardized representation of each concept, detailing key teaching points and corresponding video prompts. 
\greenCircledNumber{4} Example concepts with teaching points, visual prompts, and representative generated video frames. 
As shown, current T2V models often fail to produce videos that are both semantically aligned and physically plausible—for example, in T04 (Relative Motion), the two toy cars were intended to move side by side at the same speed, but the generated video deviates from this.}\label{fig:intro}
\vspace{-0.4cm}
\end{figure*}

\section{Introduction}
\label{sec:intro}

Creating educational videos is a resource-intensive task that requires crafting clear explanations, designing effective visuals, and ensuring both accuracy and engagement. In subjects such as physics, videos are particularly powerful, as they can vividly illustrate abstract ideas—such as motion, force, or energy—that are otherwise difficult to convey through text alone.

In recent years, there has been growing interest in leveraging AI for educational content creation, ranging from generating textual explanations to building interactive tutors and, more recently, developing multimodal learning resources~\cite{WANG2024124167,Heilala_2025,Bewersdorff_2025,yaacoub2025enhancingaidriveneducationintegrating}. Initiatives such as Khan Academy’s integration with GPT-4~\cite{khanacademyAI2024} and Socratic by Google~\cite{socraticHelpCenter} exemplify the promise of AI-powered tutoring, though they remain largely focused on text-based assistance rather than video generation. Similarly, research in intelligent tutoring systems (ITS) has advanced adaptive instruction and personalized feedback, but predominantly within textual or structured interaction formats.

Meanwhile, recent progress in text-to-video (T2V) models~\cite{chen2023videocrafter1opendiffusionmodels,chen2024videocrafter2overcomingdatalimitations,hong2022cogvideolargescalepretrainingtexttovideo,yang2025cogvideoxtexttovideodiffusionmodels,sun2024hunyuanlargeopensourcemoemodel,wan2025wanopenadvancedlargescale,wang2025klingfoleymultimodaldiffusiontransformer,liu2024sorareviewbackgroundtechnology,wang2024pikaempoweringnonprogrammersauthor,bartal2024lumierespacetimediffusionmodel} offers the potential to automatically generate rich visual explanations from natural language prompts. While these models can already produce aesthetically compelling videos, their educational utility—particularly in physics—remains underexplored~\cite{xue2025phyt2vllmguidediterativeselfrefinement,yuan2023physdiffphysicsguidedhumanmotion}. Harnessing them for instructional purposes could substantially reduce the effort required to produce high-quality learning resources, while also broadening access to scientifically accurate educational content.

To advance this vision, we introduce the first benchmark specifically designed to evaluate the capacity of T2V models to generate videos that explain physics concepts in pedagogically meaningful ways. Unlike existing benchmarks~\cite{huang2023vbenchcomprehensivebenchmarksuite,huang2024vbenchcomprehensiveversatilebenchmark,zheng2025vbench20advancingvideogeneration,sun2025t2vcompbenchcomprehensivebenchmarkcompositional,bansal2024videophyevaluatingphysicalcommonsense,bansal2025videophy2challengingactioncentricphysical,meng2024worldsimulatorcraftingphysical,motamed2025generative}, which emphasize general video quality or physical plausibility, our benchmark prioritizes educational utility by grounding evaluation in well-defined physics concepts and their associated teaching points. Each concept is systematically decomposed into a set of teaching points that mirror how the concept would be introduced in instructional practice, ensuring both comprehensive coverage and pedagogical coherence. This structured design allows us to evaluate whether generated videos meaningfully support conceptual understanding rather than merely displaying visual plausibility. Figure~\ref{fig:intro} provides an overview of our benchmark.
The PhyEduVideo benchmark consists of 205 prompts spanning 60 physics concepts, each decomposed into 1–5 teaching points that directly align with instructional goals. Breaking concepts down into teaching points ensures comprehensive coverage. The prompt associated with each teaching point has an average length of 16–45 words. Among the models we analyzed, Wan2.1 achieves the strongest overall performance, followed by PhyT2V. Domains such as Mechanics, Fluids, and Optics show relatively higher accuracy, whereas Electromagnetism and Thermodynamics remain more challenging, highlighting areas for future improvement.
Our contributions are threefold:
\begin{itemize}
\item We introduce PhyEduVideo, the first physics education benchmark designed to evaluate T2V generative models.
\item We provide a structured framework that grounds evaluation in pedagogical units of analysis (teaching points), enabling fine-grained assessment of educational utility.
\item We present empirical insights into the strengths and limitations of current T2V models in generating instructional videos, showing that while they produce visually coherent outputs, they often struggle with physics commonsense and semantic alignment.
\end{itemize}

\section{Related Work}
\label{sec:rw}

\begin{figure*}
  \centering
  \begin{subfigure}[b]{0.32\textwidth}
    \includegraphics[width=\textwidth]{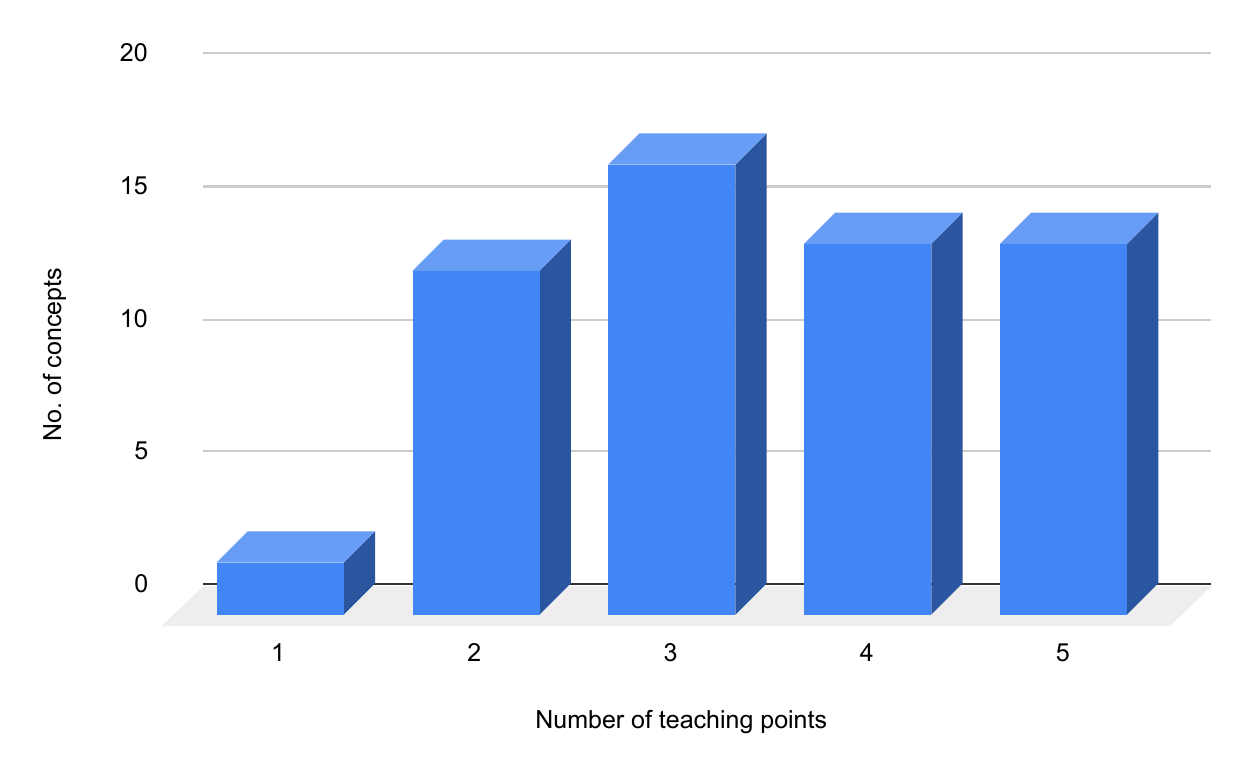}
    \caption*{(a)}
  \end{subfigure}
  \hfill
  \begin{subfigure}[b]{0.32\textwidth}
    \includegraphics[width=\textwidth]{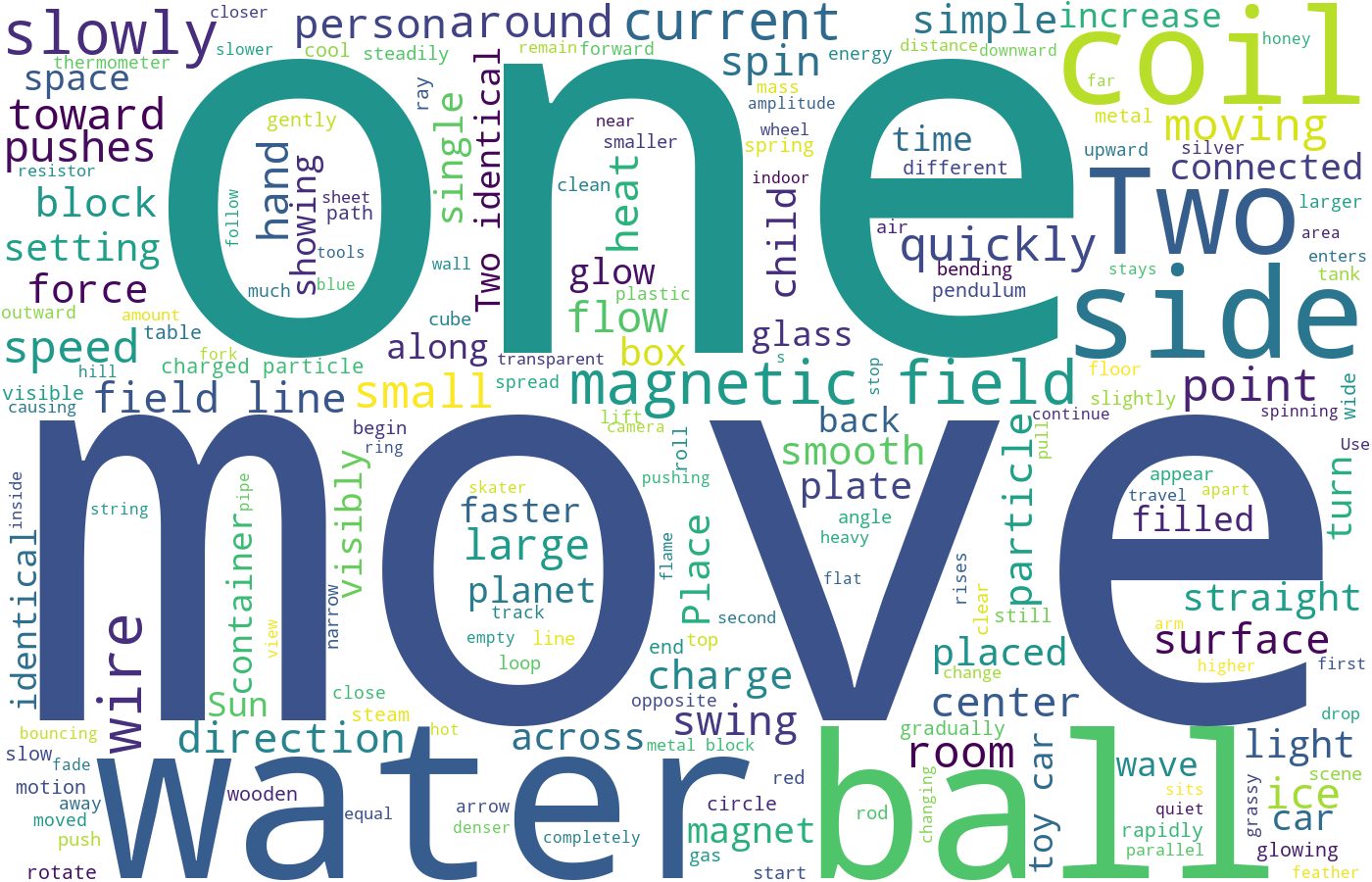}
    \caption*{(b)}
  \end{subfigure}
  \hfill
  \begin{subfigure}[b]{0.32\textwidth}
    \includegraphics[width=\textwidth]{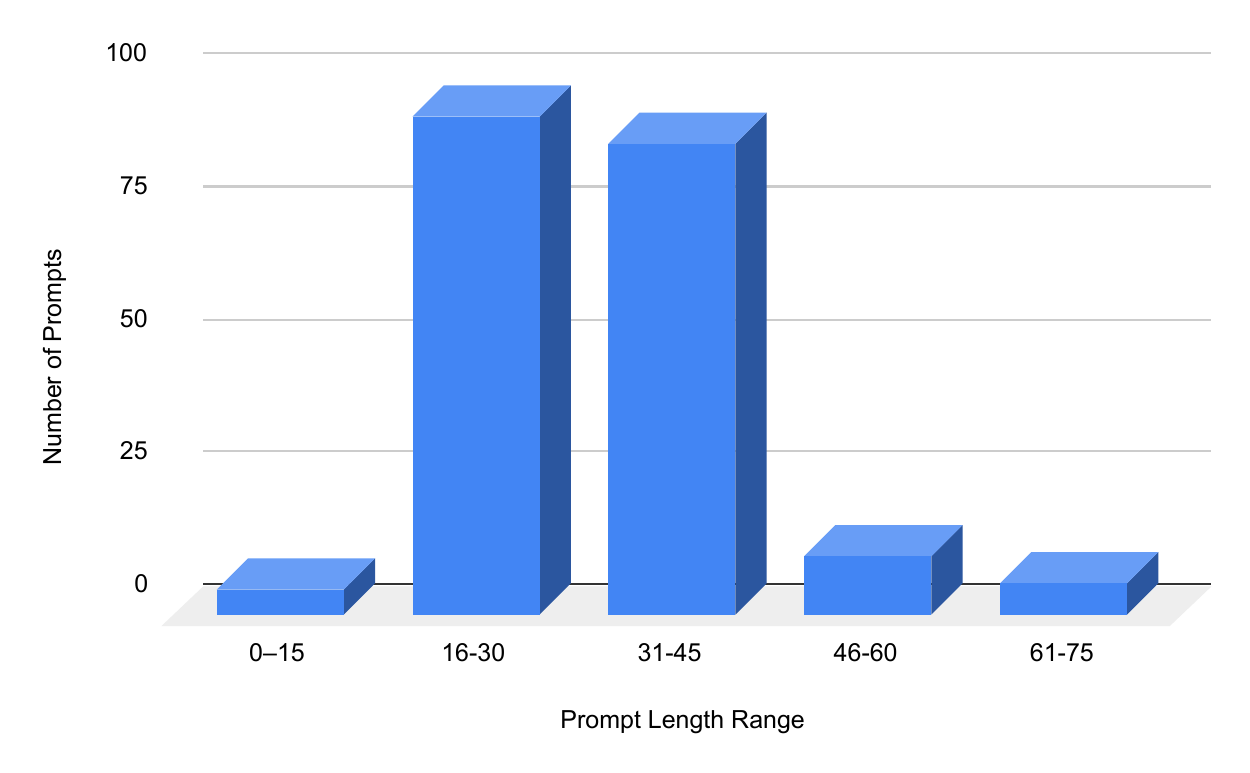}
    \caption*{(c)}
  \end{subfigure}
  \caption{\textbf{Overview of benchmark statistics for the PhyEduVideo dataset.} (a) Distribution of teaching points across physics concepts, (b) Word cloud of frequent prompt terms, (c) Distribution of prompt lengths.}
  \label{fig:phyeduvideo_stats}
  \vspace{-0.5cm}
\end{figure*}

\subsection{Text-to-Video Models}

Text-to-video (T2V) generation has advanced rapidly, evolving from early GAN-based systems to diffusion and transformer architectures. Initial approaches such as MoCoGAN~\cite{tulyakov2017mocogandecomposingmotioncontent} and TGAN~\cite{ding2019tgandeeptensorgenerative} introduced spatiotemporal discriminators but suffered from poor scalability, motion consistency, and text alignment. Diffusion models soon became dominant, with UNet-based architectures progressively denoising latent representations into coherent frames. Representative examples include ModelScope~\cite{wang2023modelscopetexttovideotechnicalreport}, VideoCrafter~\cite{chen2023videocrafter1opendiffusionmodels,chen2024videocrafter2overcomingdatalimitations}, CogVideo~\cite{hong2022cogvideolargescalepretrainingtexttovideo}, AnimateDiff~\cite{guo2024animatediffanimatepersonalizedtexttoimage}, and Text2Video-Zero~\cite{khachatryan2023text2videozerotexttoimagediffusionmodels}. Large-scale efforts such as Imagen Video~\cite{ho2022imagenvideohighdefinition} and Make-A-Video~\cite{singer2022makeavideotexttovideogenerationtextvideo} demonstrated high-resolution synthesis and spurred widespread adoption.

However, convolutional UNets struggle with long-range temporal dependencies, motivating the shift to Diffusion Transformers (DiTs), which use self-attention to model global spatial-temporal relationships. Models such as Sora~\cite{liu2024sorareviewbackgroundtechnology}, CogVideoX~\cite{yang2025cogvideoxtexttovideodiffusionmodels}, Hunyuan~\cite{sun2024hunyuanlargeopensourcemoemodel}, Wan2.1~\cite{wan2025wanopenadvancedlargescale}, Pika~\cite{wang2024pikaempoweringnonprogrammersauthor}, Lumiere~\cite{bartal2024lumierespacetimediffusionmodel}, and Kling~\cite{wang2025klingfoleymultimodaldiffusiontransformer} exemplify this trend, setting a new paradigm for T2V. This progression from GANs to UNet diffusion, and now to transformer-driven architectures, reflects a broader shift in generative modeling toward scalability, semantic fidelity, and controllability.

Complementary research explores multimodal pretraining for scalable video understanding, retrieval-augmented generation for stronger text–video alignment, and physics-aware conditioning for controllable dynamics. Notably, PhyT2V~\cite{xue2025phyt2vllmguidediterativeselfrefinement} combines LLM guidance with simulation priors, achieving $2.3\times$ stronger physical compliance and 35\% average gains on PhyGenBench~\cite{meng2024worldsimulatorcraftingphysical}. These developments signal the maturation of T2V into a discipline uniting vision, language, and physical reasoning. Physics-aware models, in particular, show promise for education by offering intuitive, visual explanations of abstract concepts. By explicitly simulating physical interactions and constraints, they open new opportunities for delivering pedagogically grounded resources at scale. In this study, we systematically evaluate their strengths and limitations for instructional use.

\subsection{Evaluation Benchmarks for Text-to-Video Models}

While T2V models have made rapid progress in fidelity, stability, and semantic alignment, their evaluation has relied mostly on general-purpose metrics. VBench~\cite{huang2023vbenchcomprehensivebenchmarksuite} introduced a hierarchical framework with dimensions such as prompt adherence, spatial coherence, and temporal consistency, later expanded in VBench++~\cite{huang2024vbenchcomprehensiveversatilebenchmark} and VBench 2.0~\cite{zheng2025vbench20advancingvideogeneration} to include commonsense reasoning, physics realism, and aesthetics. Other benchmarks focus on compositional generalization (T2VCompBench), motion dynamics (DEVIL~\cite{liao2024evaluationtexttovideogenerationmodels}), or controllability. Together, these efforts have established a solid foundation for large-scale and systematic T2V evaluation.

Physics-specific benchmarks test adherence to physical principles. VideoPhy~\cite{bansal2024videophyevaluatingphysicalcommonsense} introduced Semantic Adherence (SA) and Physical Commonsense (PC) metrics, extended in VideoPhy2~\cite{bansal2025videophy2challengingactioncentricphysical} with a Physical Rules (PR) dimension. Physics-IQ~\cite{motamed2025generative} emphasized intuitive physical reasoning with real-world videos, while PhyGenBench~\cite{meng2024worldsimulatorcraftingphysical} broadened coverage across mechanics, thermodynamics, and optics using simulation probes and LLM-based evaluators. These benchmarks represent an important step toward measuring physical realism, yet they are not explicitly designed for teaching contexts.

Despite these advances, existing benchmarks emphasize plausibility over pedagogy: they test if videos look realistic but not whether they \emph{teach}. To address this gap, we propose the first benchmark tailored to physics education. Each concept is decomposed into fine-grained teaching points, enabling systematic evaluation of whether generated videos convey core ideas clearly and coherently. This reframing shifts evaluation from surface-level realism to instructional utility, offering a complementary perspective to prior benchmarks and advancing T2V research toward impactful educational applications.

\section{PhyEduVideo}
\label{sec:phyedu}
The PhyEduVideo benchmark is developed to systematically evaluate the capabilities of Text-to-Video (T2V) models in accurately visualizing foundational physics concepts for educational purposes. It encompasses a total of 60 core concepts drawn from seven major domains of classical physics: \textit{Mechanics} (38.33\%), \textit{Waves \& Oscillations} (6.67\%), \textit{Thermodynamics} (16.67\%), \textit{Electricity and Magnetism (Electromagnetism)} (20.00\%), \textit{Fluids} (10\%), and \textit{Optics} (8.33\%) Figure~\ref{fig:intro} \greenCircledNumber{2}. \textit{Mechanics} is the most represented domain, reflecting its foundational role in introductory physics education. A standardized format is provided in Figure~\ref{fig:intro}\greenCircledNumber{3}.

To construct the benchmark, we followed a structured multi-stage pipeline, visualized in Figure~\ref{fig:intro} \greenCircledNumber{1}:

\begin{enumerate}
\item \textbf{Concept Identification:} We began by using the Gemma-3 language model~\cite{team2025gemma} to extract an initial set of classical physics concepts from standard K-12 and undergraduate physics curricula. This automated step ensured coverage across a wide conceptual space.

\item \textbf{Manual Filtering:} The extracted list was then manually reviewed by physics experts to retain only those concepts that are both pedagogically essential and visually realizable. Abstract, redundant, or highly mathematical topics—such as Lagrangian mechanics, tensor calculus, or complex integrals—were excluded in favor of those that lend themselves to intuitive, observable phenomena like Newton’s laws, simple harmonic motion, or conservation of energy.

\item \textbf{Decomposition into Teaching Points:} Each validated concept was further broken down into multiple \textit{teaching points}—fine-grained, pedagogically distinct sub-concepts that capture specific physical behaviors or relationships. As shown in Figure~\ref{fig:intro} \greenCircledNumber{4} for example, the concept of “Newton’s First Law” is divided into four teaching points: objects at rest, constant motion, inertia, and force-induced acceleration. This decomposition allows T2V models to be tested on precise subcomponents of conceptual understanding, rather than broad themes.

\item \textbf{Prompt Generation and Refinement:} For each teaching point, candidate prompts were first generated automatically using Gemma-3 and then refined by humans. These prompts provide short, clear descriptions for generating videos. Examples of the final prompts and their corresponding videos are shown in Figure~\ref{fig:intro} \greenCircledNumber 4.
\end{enumerate}

\textbf{Benchmark Statistics:}
The final PhyEduVideo benchmark comprises 205 prompts derived from the 60 physics concepts, each decomposed into between one and five teaching points (Figure~\ref{fig:phyeduvideo_stats}(a)). Each prompt is written as a self-contained, visually descriptive scenario that maps directly to a teaching goal. The average prompt length falls in the 16–45 word range, with longer prompts offering additional context for more complex situations, as seen in Figure~\ref{fig:phyeduvideo_stats}(c). Figure~\ref{fig:phyeduvideo_stats}(b) shows the prompt vocabulary, which spans a wide range of physical entities (e.g., “ball,” “coil,” “current”) and actions (e.g., “move,” “push,” “show”), reflecting both linguistic diversity and conceptual coverage. Collectively, these characteristics enable PhyEduVideo to serve as a rigorous and pedagogically grounded testbed for evaluating the scientific accuracy, temporal coherence, and visual fidelity of physics-focused T2V models.
In comparison, PhyGenBench offers 160 prompts across 27 physical laws, T2VPhysBench provides 84 prompts spanning twelve laws, and VideoPhy focuses on interaction-driven scenarios—highlighting PhyEduVideo’s broader, education-oriented design grounded in structured teaching points.

\begin{table*}[ht]
\centering
\renewcommand{\arraystretch}{1.15}
\setlength{\tabcolsep}{5pt}

\begin{tabular}{l cc|cc|cc|cc|cc|cc|cc}
\toprule
Metric 
  & \multicolumn{2}{c}{\textit{EM}} 
  & \multicolumn{2}{c}{\textit{Mech}} 
  & \multicolumn{2}{c}{\textit{Fluids}} 
  & \multicolumn{2}{c}{\textit{Thermal}} 
  & \multicolumn{2}{c}{\textit{Optics}} 
  & \multicolumn{2}{c}{\textit{W\&O}} 
  & \multicolumn{2}{c}{Avg} \\

\cmidrule(lr){2-3}
\cmidrule(lr){4-5}
\cmidrule(lr){6-7}
\cmidrule(lr){8-9}
\cmidrule(lr){10-11}
\cmidrule(lr){12-13}
\cmidrule(lr){14-15}

& $\rho \uparrow$ & $\tau \uparrow$
  & $\rho \uparrow$ & $\tau \uparrow$
  & $\rho \uparrow$ & $\tau \uparrow$
  & $\rho \uparrow$ & $\tau \uparrow$
  & $\rho \uparrow$ & $\tau \uparrow$
  & $\rho \uparrow$ & $\tau \uparrow$
  & $\rho \uparrow$ & $\tau \uparrow$ \\

\midrule

VideoPhy-SA   & 0.24 & 0.19 & 0.31 & 0.24 & 0.49 & 0.40 & 0.28 & 0.21 & 0.45 & 0.36 & 0.47 & 0.37 & 0.44 & 0.34 \\
VideoPhy-PC   & -0.01 & -0.01 & 0.11 & 0.09 & -0.11 & -0.09 & -0.05 & -0.04 & 0.17 & 0.14 & 0.07 & 0.06 & 0.01 & 0.01 \\
PhyEduVideo-SA & 0.46 & 0.41 & 0.48 & 0.45 & 0.59 & 0.51 & 0.66 & 0.60 & 0.45 & 0.41 & 0.42 & 0.39 & 0.51 & 0.46 \\
PhyEduVideo-PC & 0.30 & 0.27 & 0.56 & 0.52 & 0.35 & 0.33 & 0.59 & 0.55 & 0.30 & 0.27 & 0.57 & 0.54 & 0.39 & 0.36 \\

\bottomrule
\end{tabular}

\caption{
Domain-wise correlations between human and model scores using Spearman’s $\rho$ and Kendall’s $\tau$. Models (VideoPhy, PhyEduVideo) are split into SA = Semantic Alignment and PC = Physics Commonsense. 
Domains are abbreviated as follows: 
EM: Electromagnetism, Mech: Mechanics,
Thermal: Thermodynamics, W\&O: Waves and Oscillations, 
and Avg: Average across all domains. 
}
\label{tab:cat-wise-correl}
\vspace{-3mm}
\end{table*}

\begin{figure*}[h]
    \centering
    \includegraphics[width=\textwidth]{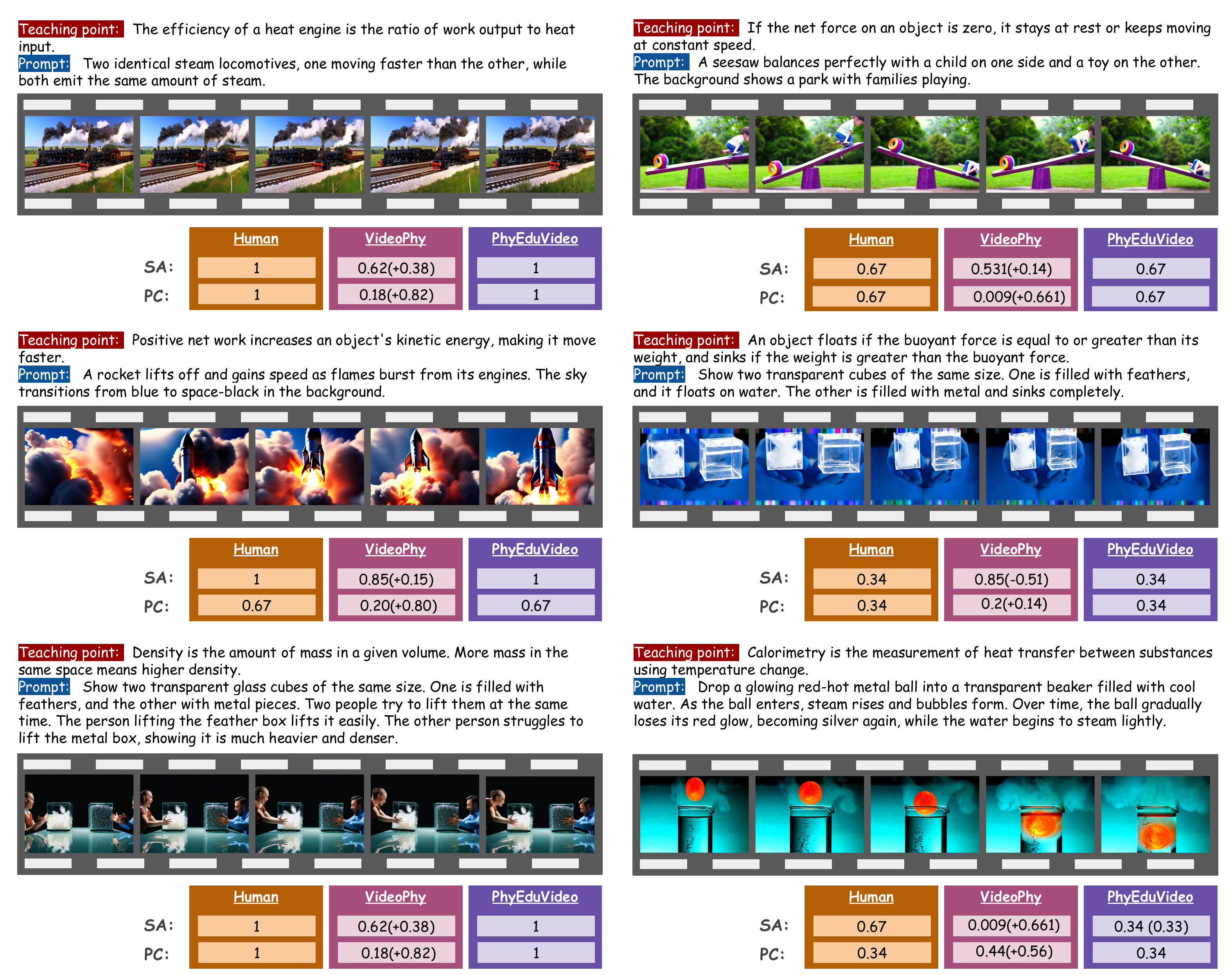}
    \caption{Comparison of SA (Semantic Adherence) and PC (Physics Commonsense) scores assigned by the VideoPhy, Automatic Evaluator (PhyEduVideo) and humans. Detailed videos are available on the GitHub page.}
    \label{fig:corr_qualitative_results}
\vspace{-6mm}
\end{figure*}

\subsection{Metrics}
\label{sec:eval}
To evaluate T2V models for physics education using the PhyEduVideo benchmark, we propose a structured framework assessing video generation quality, prompt adherence, and physics-specific fidelity. The evaluation is conducted across four dimensions: \textit{Semantic Alignment (SA)}, \textit{Physical Commonsense (PC)}, \textit{Motion Smoothness (MS)}, and \textit{Temporal Flickering (TF)}.

\begin{itemize}
    \item \textbf{Semantic Alignment (SA):}~\cite{bansal2024videophyevaluatingphysicalcommonsense,bansal2025videophy2challengingactioncentricphysical,meng2024worldsimulatorcraftingphysical} This metric measures how well a generated video matches the main idea of the input prompt. It checks if the core scenario, key actions, and important visual elements described in the text appear correctly and coherently in the video. For example, for the prompt ``A rolling ball continues in a straight line,'' a semantically aligned video should show the ball moving steadily along a straight path. Semantic Alignment is scored from 0 to 3 using InternVL3.5~\cite{wang2025internvl35advancingopensourcemultimodal}, which evaluates two components: object score (0 = none, 1 = some, 2 = all key objects present) and action score (0 = main action not depicted, 1 = main action depicted). A higher score means the video correctly represents both the described objects and actions.

   \item \textbf{Physics Commonsense (PC):}~\cite{bansal2024videophyevaluatingphysicalcommonsense,bansal2025videophy2challengingactioncentricphysical,meng2024worldsimulatorcraftingphysical}
This metric evaluates whether the generated video correctly follows the intended teaching point. For example, when ice is placed in water, it should melt gradually, \(0^{\circ}C\) until the ice is fully melted, and the water level should rise steadily as the ice turns into liquid.
Following PhyGenBench~\cite{bansal2024videophyevaluatingphysicalcommonsense}, this metric is structured into three finer-grained evaluation stages:
\begin{enumerate}
    \item \textit{Key Physical Phenomena Detection:} This sub-metric evaluates whether the video successfully captures the essential physical behavior described in the prompt. For example, if the prompt involves projectile motion, the video should display a curved parabolic trajectory, rather than an unrealistic linear path.

    \item \textit{Physics Order Verification:} This stage assesses the temporal coherence of physical events within the video. It verifies whether the sequence of actions follows a logically and physically correct order. For instance, in a pendulum motion, the object must first be released before it begins to swing. To perform this evaluation automatically, we employ LLaVA-Interleave~\cite{li2024llavanextinterleavetacklingmultiimagevideo}.

    \item \textit{Overall Naturalness Evaluation:} This component assesses the naturalness of a video by examining whether objects and their movements appear physically plausible. To guide this evaluation, we define four GPT-generated descriptions for a given prompt, representing different levels of naturalness: Fantastical descriptions involve highly imaginative or impossible scenarios. Clearly unrealistic descriptions depict objects behaving in ways that blatantly violate fundamental physical principles, for example, a ball sinking through a solid table or two objects occupying the same space simultaneously. Slightly unrealistic descriptions generally follow physical principles but include minor inconsistencies or exaggerated effects, such as overly bouncy objects or frictionless slides. Realistic descriptions describe objects moving and interacting fully in accordance with real-world physics. InternVideo2~\cite{wang2024internvideo2scalingfoundationmodels} is then employed to compare the generated video against these categories, assigning the most appropriate category to the video.
\end{enumerate}

   \item \textbf{Motion Smoothness:}~\cite{huang2023vbenchcomprehensivebenchmarksuite} This refers to the continuity and coherence of object motion and background in the video. Videos should not exhibit jerky, inconsistent, or mechanically impossible motion patterns. The motion in the video should be smooth and follow the physics concept.

    \item \textbf{Temporal Flickering:}~\cite{huang2023vbenchcomprehensivebenchmarksuite} This evaluates the stability of visual properties (like object color, size, or shape) across frames. Abrupt flickers, changes in object identity, or disappearing elements can break temporal coherence and degrade the viewing experience. A consistently rendered object across the video receives a high flickering score.
\end{itemize}

Overall, these four criteria provide a structured and holistic framework for evaluating generated videos in the context of physics education. By addressing both conceptual and visual aspects, the benchmark supports rigorous and pedagogically meaningful assessment of model outputs. This enables more targeted progress in developing text-to-video models that are both scientifically accurate and educationally effective.

\subsection{Human Evaluation}
\label{sec:human_evals}

To assess the alignment of automatic metrics with human perception, we conducted a human evaluation study on 500 videos, involving annotators who had formally studied physics up to the 12\textsuperscript{th} grade. The results, summarized in Tables~\ref{tab:cat-wise-correl}, show that \textbf{PhyEduVideo} achieves much stronger correlations with human judgments than \textbf{VideoPhy}~\cite{bansal2024videophyevaluatingphysicalcommonsense}. VideoPhy is a benchmark that tests whether text-to-video models follow basic physical commonsense, such as correct object interactions, material behaviors, and physical laws along with semantic adherence. For both SA and PC, the highest correlations are observed in the \textit{Thermodynamics} category, while the lowest are found in \textit{Electromagnetism} and \textit{Optics}. Overall, PhyEduVideo achieves a Spearman correlation of 0.509 and a Kendall correlation of 0.462 for SA---considerably higher than the corresponding values for VideoPhy (gap = 0.071 and 0.122). For PC, PhyEduVideo reaches 0.392 (Spearman) and 0.363 (Kendall), again significantly outperforming VideoPhy (0.008 and 0.006), with absolute gains of 0.384 and 0.357, respectively. This consistent gap underscores the value of our benchmark in better capturing human judgment.  Importantly, these higher correlation numbers also indicate that \textbf{PhyEduVideo} more faithfully aligns with pedagogically accurate teaching points, ensuring that evaluation outcomes reflect not just visual plausibility but instructional relevance. Figure~\ref{fig:corr_qualitative_results} presents qualitative examples where human scores are shown alongside predictions from VideoPhy and PhyEduVideo, further demonstrating how our benchmark provides more faithful and interpretable assessments.

\begin{table}[t]
\centering
\begin{tabular}{lcccccc}
\hline
 & Model & SA $\uparrow$ & PC $\uparrow$ & MS $\uparrow$ & TF $\uparrow$\\
\hline

\multirow{5}{*}{\rotatebox{90}{\textit{Mechanics}}} 
    & VideoCrafter2 & 0.75 & 0.52 & 0.94 & 0.92 \\
    & CogVideoX     & \cellcolor{cyan!15}0.85 & 0.57 & \cellcolor{cyan!15}0.98 & 0.97 \\
    & Wan2.1        & \cellcolor{cyan!50}0.86 & \cellcolor{cyan!50}0.66 & \cellcolor{cyan!50}0.99 & \cellcolor{cyan!15}0.98 \\
    & Video-MSG     & 0.75 & 0.53 & \cellcolor{cyan!50}0.99 & \cellcolor{cyan!50}0.99 \\
    & PhyT2V        & 0.80 & \cellcolor{cyan!15}0.59 & \cellcolor{cyan!15}0.98 & 0.97 \\
\hline

\multirow{5}{*}{\rotatebox{90}{\textit{W\&O}}} 
    & VideoCrafter2 & 0.72 & \cellcolor{cyan!15}0.49 & 0.92 & 0.90 \\
    & CogVideoX     & \cellcolor{cyan!15}0.79 & \cellcolor{cyan!50}0.59 & \cellcolor{cyan!15}0.98 & \cellcolor{cyan!15}0.98 \\
    & Wan2.1        & \cellcolor{cyan!50}0.87 & \cellcolor{cyan!50}0.59 & \cellcolor{cyan!50}0.99 & \cellcolor{cyan!15}0.98 \\
    & Video-MSG     & 0.69 & 0.46 & \cellcolor{cyan!50}0.99 & \cellcolor{cyan!50}0.99 \\
    & PhyT2V        & 0.72 & \cellcolor{cyan!50}0.59 & \cellcolor{cyan!15}0.98 & 0.97 \\
    
\hline

\multirow{5}{*}{\rotatebox{90}{\textit{Fluids}}} 
    & VideoCrafter2 & 0.58 & 0.48 & 0.89 & 0.87 \\
    & CogVideoX     & 0.71 & \cellcolor{cyan!15}0.58 & \cellcolor{cyan!15}0.98 & 0.97 \\
    & Wan2.1        & \cellcolor{cyan!50}0.90 & \cellcolor{cyan!50}0.63 & \cellcolor{cyan!50}0.99 & \cellcolor{cyan!15}0.98 \\
    & Video-MSG     & 0.67 & \cellcolor{cyan!15}0.58 & \cellcolor{cyan!50}0.99 & \cellcolor{cyan!50}0.99 \\
    & PhyT2V        & \cellcolor{cyan!15}0.85 & \cellcolor{cyan!50}0.63 & \cellcolor{cyan!15}0.98 & 0.97 \\    
\hline

\multirow{5}{*}{\rotatebox{90}{\textit{Thermal}}}
    & VideoCrafter2 & 0.51 & 0.38 & 0.89 & 0.86 \\
    & CogVideoX     & \cellcolor{cyan!15}0.75 & \cellcolor{cyan!50}0.52 & \cellcolor{cyan!15}0.98 & \cellcolor{cyan!15}0.98 \\
    & Wan2.1        & \cellcolor{cyan!50}0.93 & \cellcolor{cyan!50}0.52 & \cellcolor{cyan!50}0.99 & \cellcolor{cyan!15}0.98 \\
    & Video-MSG     & 0.71 & 0.39 & \cellcolor{cyan!50}0.99 & \cellcolor{cyan!50}0.99 \\
    & PhyT2V        & \cellcolor{cyan!15}0.75 & \cellcolor{cyan!15}0.49 & \cellcolor{cyan!15}0.98 & 0.97 \\
\hline

\multirow{5}{*}{\rotatebox{90}{\textit{EM}}}
    & VideoCrafter2 & 0.54 & 0.50 & 0.89 & 0.88 \\
    & CogVideoX     & \cellcolor{cyan!15}0.73 & \cellcolor{cyan!50}0.65 & \cellcolor{cyan!15}0.98 & \cellcolor{cyan!15}0.98 \\
    & Wan2.1        & 0.65 & 0.57 & \cellcolor{cyan!50}0.99 & \cellcolor{cyan!15}0.98 \\
    & Video-MSG     & 0.60 & 0.48 & \cellcolor{cyan!50}0.99 & \cellcolor{cyan!50}0.99 \\
    & PhyT2V        & \cellcolor{cyan!50}0.75 & \cellcolor{cyan!15}0.62 & \cellcolor{cyan!15}0.98 & \cellcolor{cyan!15}0.98 \\
\hline

\multirow{5}{*}{\rotatebox{90}{\textit{Optics}}} 
    & VideoCrafter2 & 0.64 & 0.62 & \cellcolor{cyan!15}0.88 & 0.83 \\
    & CogVideoX     & 0.69 & 0.60 & \cellcolor{cyan!50}0.99 & \cellcolor{cyan!15}0.98 \\
    & Wan2.1        & \cellcolor{cyan!15}0.78 & \cellcolor{cyan!15}0.64 & \cellcolor{cyan!50}0.99 & \cellcolor{cyan!15}0.98 \\
    & Video-MSG     & 0.69 & \cellcolor{cyan!15}0.64 & \cellcolor{cyan!50}0.99 & \cellcolor{cyan!50}0.99 \\
    & PhyT2V        & \cellcolor{cyan!50}0.80 & \cellcolor{cyan!50}0.71 & \cellcolor{cyan!50}0.99 & \cellcolor{cyan!15}0.98 \\
\hline

\multirow{5}{*}{\rotatebox{90}{\textit{Average}}} 
    & VideoCrafter2 & 0.62 & 0.50 & 0.90 & 0.88 \\
    & CogVideoX     & 0.75 & \cellcolor{cyan!15}0.59 & \cellcolor{cyan!15}0.98 & \cellcolor{cyan!15}0.98 \\
    & Wan2.1        & \cellcolor{cyan!50}0.83 & \cellcolor{cyan!50}0.60 & \cellcolor{cyan!50}0.99 & \cellcolor{cyan!15}0.98 \\
    & Video-MSG     & 0.68 & 0.52 & \cellcolor{cyan!50}0.99 & \cellcolor{cyan!50}0.99 \\
    & PhyT2V        & \cellcolor{cyan!15}0.78 & \cellcolor{cyan!50}0.60 & \cellcolor{cyan!15}0.98 & 0.97 \\
\hline
\end{tabular}
\caption{
Comparison of five video generation models across six physics domains, along with their overall averages. Metrics include Semantic Adherence (SA), Physics Commonsense (PC), Motion Smoothness (MS), and Temporal Flickering (TF). Best scores are highlighted in cyan, and second-best in light cyan. Domains are abbreviated as follows: 
EM: Electromagnetism,
Thermal: Thermodynamics and W\&O: Waves and Oscillations
}
\label{tab:cat_scores}
\vspace{-0.7cm}
\end{table}

\section{Experiments}
\subsection{Evaluated Models}
\label{sec:exp}
We evaluate five state-of-the-art text-to-video (T2V) generation models on our benchmark: CogVideoX~\cite{yang2025cogvideoxtexttovideodiffusionmodels}, Wan2.1~\cite{wan2025wanopenadvancedlargescale}, VideoCrafter2~\cite{chen2024videocrafter2overcomingdatalimitations}, Video‑MSG~\cite{li2025trainingfreeguidancetexttovideogeneration}, and PhyT2V~\cite{meng2024worldsimulatorcraftingphysical}. CogVideoX-5B, with demonstrated success on physics-focused evaluations, serves as a baseline for physics-grounded video generation due to its consistent high scores in physics-following benchmarks. Wan2.1 is a strong, general-purpose T2V model that achieves high scores across a wide range of benchmarks, providing insight into the generalization capabilities of current systems. VideoCrafter2 is known for generating high-resolution, visually coherent videos, making it useful for assessing visual quality and detail.
In addition, we include models with more specialized architectures. Video‑MSG employs a training-free, structured guidance pipeline that closely follows input prompts. Its generation proceeds in three stages: (1) Background Planning, where a multimodal large language model (MLLM, specifically GPT-4o) produces detailed background descriptions, rendered via a text-to-image (T2I) model and animated with an image-to-video (I2V) model; (2) Foreground Object Layout and Trajectory Planning, where object positions and motions are inferred with MLLM guidance; and (3) Video Generation, where the planned layout is denoised to produce the final video. This compositional approach has shown strong performance on T2V-CompBench~\cite{sun2025t2vcompbenchcomprehensivebenchmarkcompositional} and is evaluated here for its ability to produce visually coherent, pedagogically meaningful physics content.
PhyT2V~\cite{xue2025phyt2vllmguidediterativeselfrefinement}, in contrast, is specifically engineered for physics-aware generation: it integrates large language models with physics simulation priors to iteratively refine video content, ensuring adherence to physical laws while maintaining semantic and temporal coherence. Its performance on PhyGenBench~\cite{meng2024worldsimulatorcraftingphysical} demonstrates notable gains in physical plausibility and instructional clarity, making it uniquely suited for evaluating T2V models in educational contexts. 
Comprehensive implementation details, including model configurations and evaluation protocols, are provided in the Appendix~\ref{model details}.
\subsection{Quantitative Evaluation}
\label{sec:quantitative}

Quantitative evaluations, summarized in Tables~\ref{tab:cat_scores}, highlight clear trends in both perceptual quality and correctness-based performance of text-to-video (T2V) models. All evaluated models achieve consistently high scores in Motion Smoothness (MS) and Temporal Flickering (TF), with values typically above 0.85, demonstrating that current systems are capable of generating visually coherent and temporally stable videos. However, this strength contrasts sharply with the lower scores observed in correctness-oriented metrics such as Semantic Adherence (SA) and Physics Commonsense (PC), which are critical for ensuring educational and conceptually accurate content. Among the models, Wan2.1~\cite{wan2025wanopenadvancedlargescale} stands out as the overall best performer, achieving the highest SA and PC scores across most domains, followed closely by PhyT2V~\cite{xue2025phyt2vllmguidediterativeselfrefinement}, which maintains competitive reasoning ability while delivering visually stable results. In comparison, VideoCrafter2~\cite{chen2024videocrafter2overcomingdatalimitations} ranks lowest in both SA and PC despite its strong performance on temporal flickering and motion smoothness. Video-MSG~\cite{li2025trainingfreeguidancetexttovideogeneration} similarly excels in video quality metrics but does not achieve a significant boost in physics commonsense, suggesting that compositional control alone is insufficient for capturing complex physics concepts.

A category-level analysis reveals notable differences across physics domains. \textit{Mechanics} emerges as a relatively solvable domain, with models achieving SA scores above 0.75 and PC scores exceeding 0.50, reflecting that visually grounded concepts like motion and collisions are easier to represent. \textit{Fluids} and \textit{Optics} stand out as the best-performing domains overall (across all 4 metrics), reaching the highest SA (up to 0.90) and PC (up to 0.71), indicating that distinctive visual dynamics such as flow patterns or light interactions are more learnable by current models. By contrast, \textit{Thermodynamics} and \textit{Electromagnetism} show the weakest correctness performance: in Thermodynamics, VideoCrafter2 drops to an SA of 0.51 and a PC of 0.38, while in Electromagnetism, most models record PC values below 0.50. \textit{Waves \& Oscillations} show moderate performance, better than Thermodynamics and Electromagnetism but trailing behind Fluids and Optics. These results reveal a consistent reasoning–perception gap: while models reliably generate smooth and visually appealing content, their semantic adherence and physics commonsense remain limited. Wan2.1 and PhyT2V perform comparatively better, showing greater stability, coherence, and conceptual alignment, making them more suitable for physics-focused educational content. A key challenge arises in domains such as \textit{Electromagnetism}, where core concepts involve charges, magnetic fields, and electric fields—phenomena that are not directly visible. For teaching, however, it is crucial to make such invisible entities perceivable in order to build understanding. This is precisely where our benchmark stands out from existing physics-based benchmarks: rather than only checking whether generated videos obey physical laws, we emphasize the educational dimension, requiring models to represent abstract and invisible concepts in a way that aids learning.

\begin{figure*}[h]
    \centering
    \includegraphics[width=\textwidth,height=\textheight,keepaspectratio]{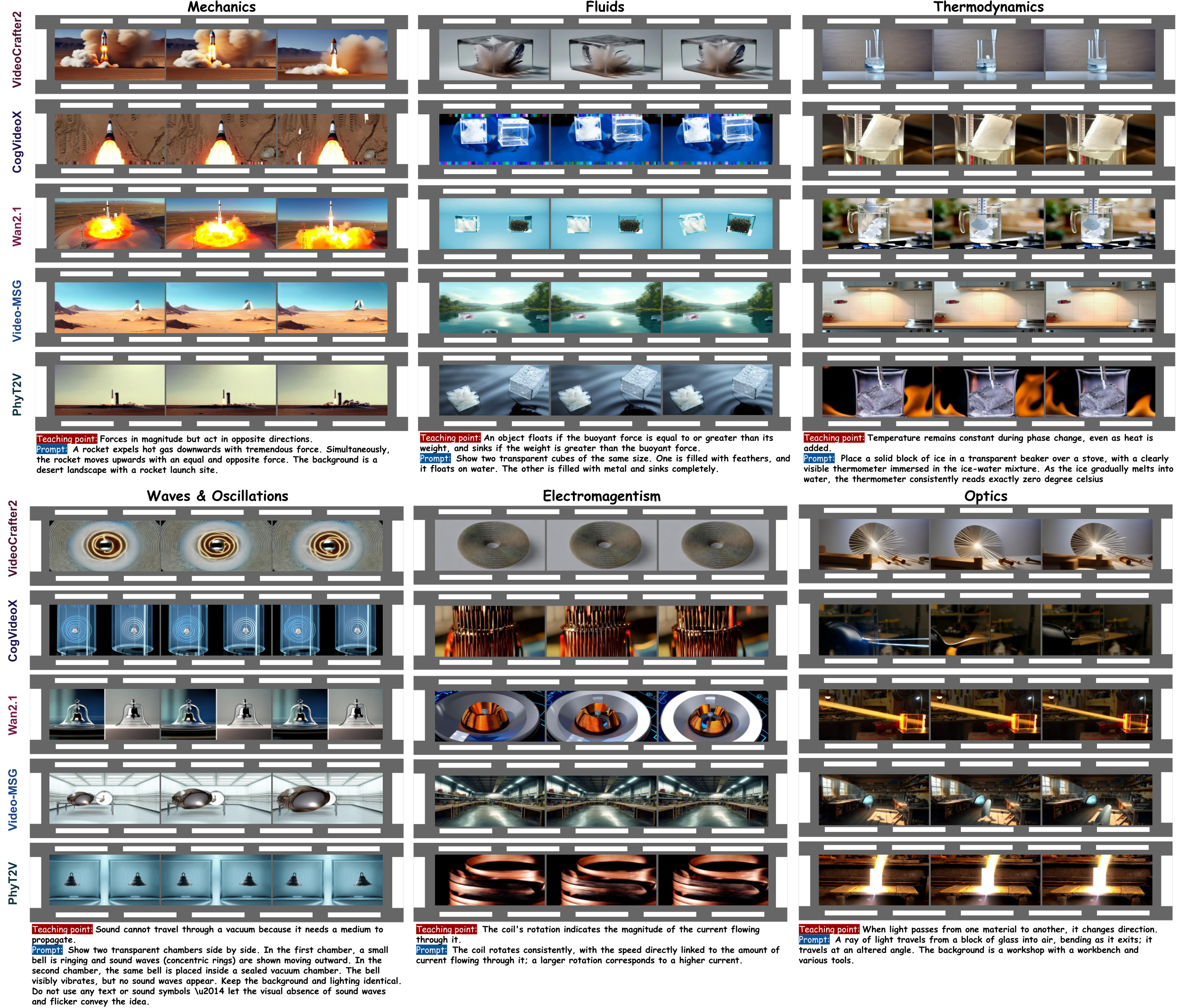}
    \vspace{-0.6cm}
    \caption{Qualitative comparisons of generated videos across six classical physics categories—\textit{Mechanics}, \textit{Waves \& Oscillations}, \textit{Fluids}, \textit{Thermodynamics}, \textit{Electromagnetism}, and \textit{Optics}—for five T2V models: VideoCrafter2, CogVideoX, Wan2.1, Video-MSG, and PhyT2V. Detailed videos are available on the GitHub page.}
    \label{fig:qualitative_results}
\vspace{-4mm}
\end{figure*}
\subsection{Qualitative Evaluation}
\label{sec:qualitive}

Figure~\ref{fig:qualitative_results} qualitatively compares generations across six key physics education domains—\textit{Mechanics}, \textit{Waves \& Oscillations}, \textit{Fluids}, \textit{Thermodynamics}, \textit{Electromagnetism}, and \textit{Optics}—revealing strengths and limitations in how current models visually communicate scientific concepts. Corresponding to each domain, an example teaching point, textual prompt to query the T2V and images from start, middle and end part of the generated output video are shown in the Figure~\ref{fig:qualitative_results}. Detailed videos are available on the GitHub pag. Wan 2.1~\cite{wan2025wanopenadvancedlargescale} shows strong educational potential, generating coherent and semantically grounded sequences that align well with physical principles, such as realistic projectile motion in Mechanics and light refraction in Optics. CogVideoX~\cite{yang2025cogvideoxtexttovideodiffusionmodels} performs well in \textit{Mechanics} and \textit{Fluids}, where object interactions are simpler and more grounded in visual cues, though often hindered by structural inconsistencies. VideoCrafter2~\cite{chen2024videocrafter2overcomingdatalimitations} consistently delivers visually smooth outputs but lacks the semantic fidelity needed for instructional clarity, especially in abstract domains like \textit{Electromagnetism} and \textit{Waves}. Video-MSG~\cite{li2025trainingfreeguidancetexttovideogeneration} maintains temporal stability and shows potential in controlled categories like Mechanics and Thermodynamics, yet struggles with conveying deeper causal relationships and dynamic variations essential for physics learning. PhyT2V~\cite{xue2025phyt2vllmguidediterativeselfrefinement}, designed with physics-awareness in mind, achieves a strong balance between visual stability and conceptual fidelity, excelling particularly in scenarios that require accurate physical reasoning, such as current-induced effects in \textit{Electromagnetism}. These observations underscore the gap between visual quality and conceptual fidelity in current models, emphasizing the need for physics-aware architectures to support meaningful and accurate science education content.

\vspace{-0.25cm}
\section{Conclusion}
\label{sec:conclusion}
\vspace{-0.15cm}

This work introduces a benchmark for evaluating text-to-video (T2V) generation in physics education. Unlike prior efforts that mainly test adherence to physical laws, our benchmark emphasizes educational relevance. Each physics concept is broken into granular teaching points, with prompts targeting their visual explanation. This enables evaluation of whether models generate videos that not only look realistic but also support teaching by making abstract or invisible entities—such as charges, fields, or wave interactions—visually understandable. Using this benchmark, we evaluate CogVideoX, Wan2.1, VideoCrafter2, Video-MSG, and PhyT2V. While models produce coherent motion with reasonable smoothness, they often struggle with semantic adherence and physics commonsense. Wan2.1 and PhyT2V perform comparatively better but still have room for improvement, highlighting the need for physics-aware, education-focused T2V systems.

\section*{Acknowledgments}
We acknowledge the support of Google Cloud credits provided through a GCP Award as part of the Gemma Academic Program.

{
    \small
    \bibliographystyle{ieeenat_fullname}
    \bibliography{main}
}

\clearpage

\FloatBarrier
\appendix
\section*{Appendix}
\section{Model Details}
\label{model details}
We evaluate six state-of-the-art video generation models with distinct design philosophies. VideoCrafter2 is an open-source diffusion-based framework known for controllability and high-quality short clips. CogVideoX~\cite{yang2025cogvideoxtexttovideodiffusionmodels}, a transformer-based model, emphasizes long-duration generation with improved temporal coherence. Wan2.1 advances photorealism and motion stability through refined denoising strategies. Video-MSG employs a controlled generation strategy getting high scores for T2VCompench~\cite{sun2025t2vcompbenchcomprehensivebenchmarkcompositional} prompts. PhyT2V~\cite{xue2025phyt2vllmguidediterativeselfrefinement} is a model designed for physics video generation via CoT method. Table~\ref{tab:model-details} represents the model details for each model.

\section{Human Evaluation}
\begin{figure*}[h]
    \centering
    \includegraphics[width=\textwidth,height=0.9\textheight,keepaspectratio]{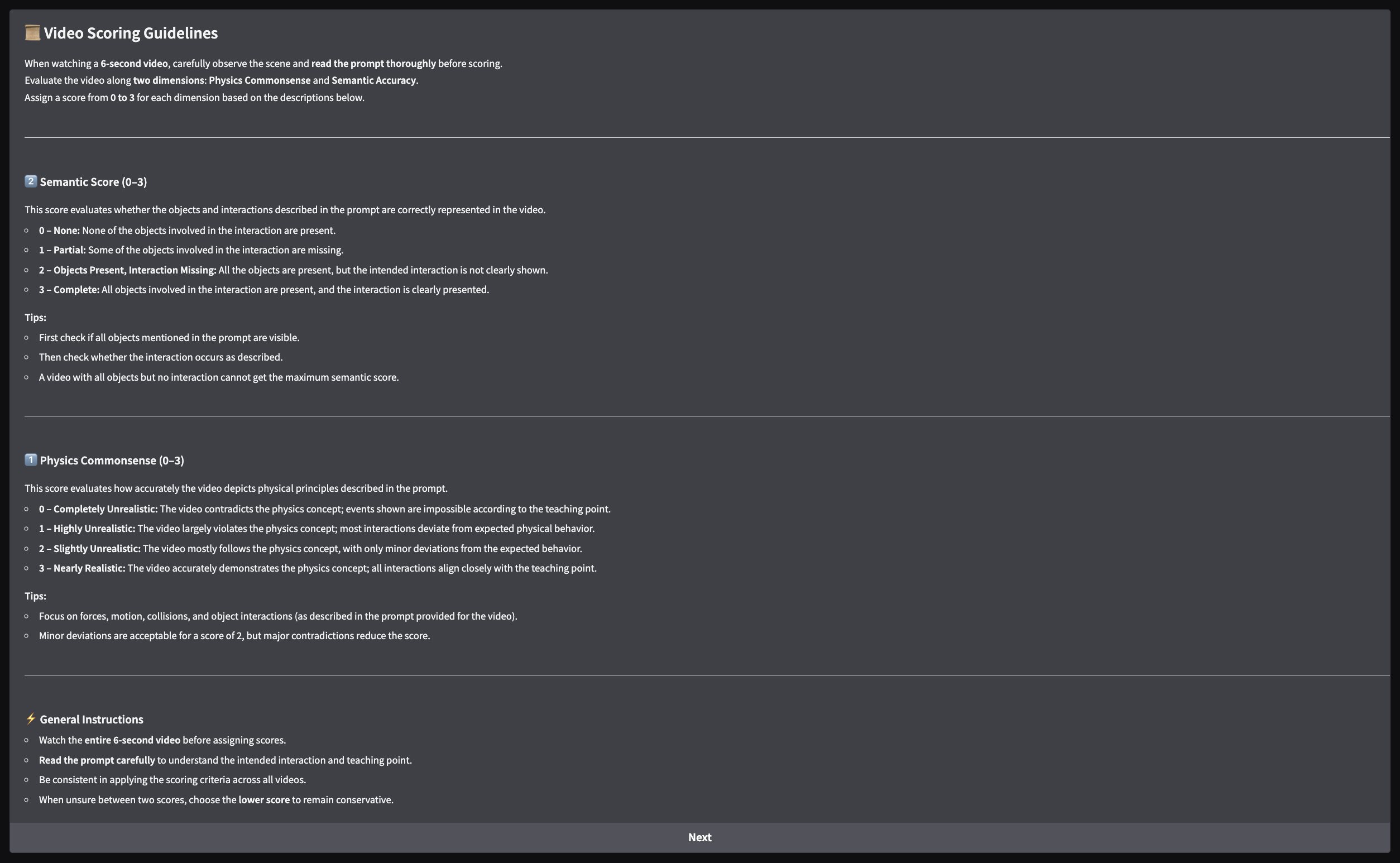}
    \caption{Guidelines and rules given to human annotators to ensure consistent and reliable evaluation.}

    \label{fig:rules}
\end{figure*}

\begin{figure*}[h]
    \centering
    \includegraphics[width=\textwidth,height=0.9\textheight,keepaspectratio]{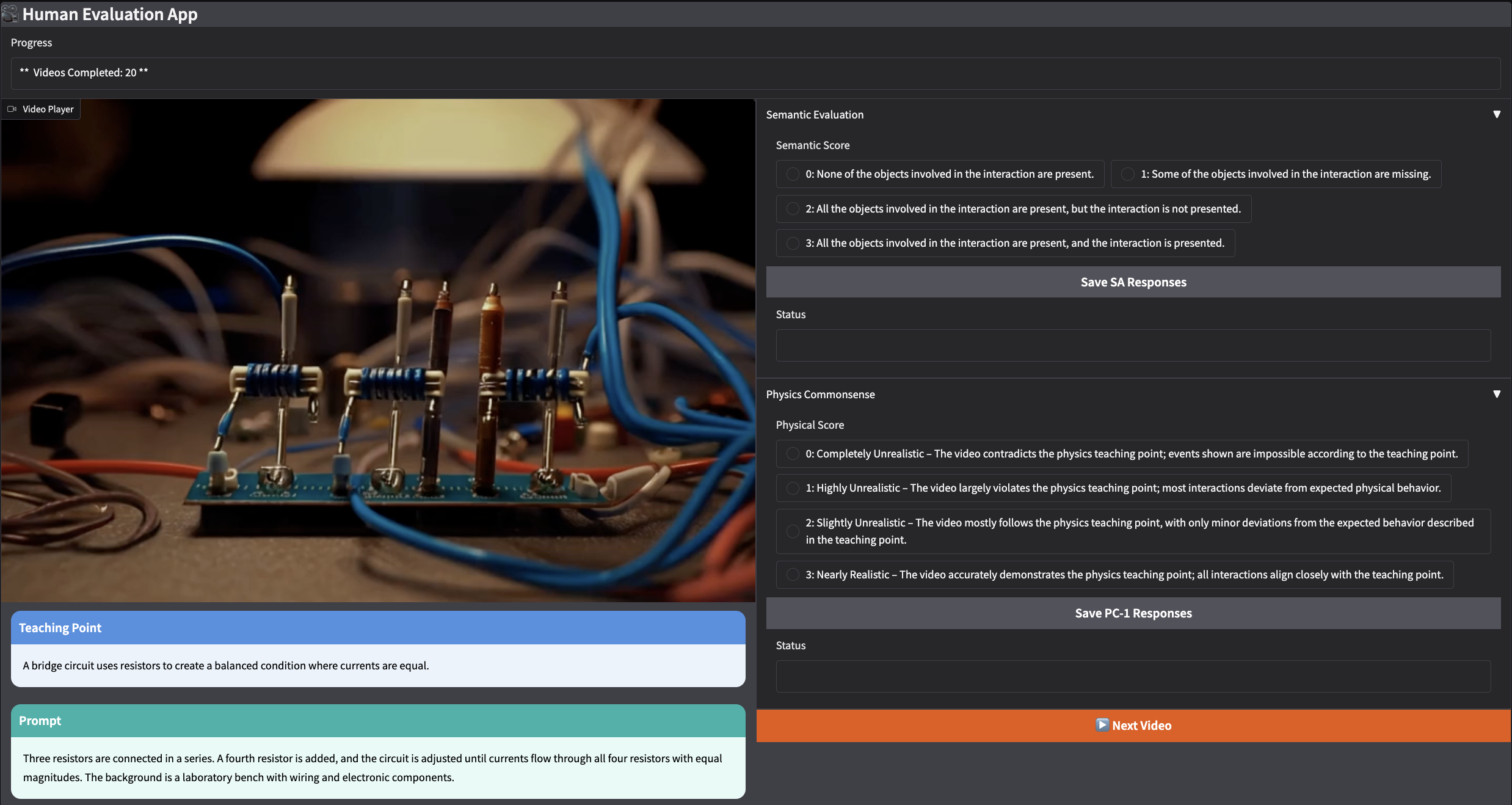}
    \caption{Questions provided for human evaluation and their respective scoring schemes are illustrated in the diagram above.}
    \label{fig:human_eval}
\end{figure*}

A total of 500 videos were selected for human evaluation, covering outputs from VideoCrafter2~\cite{chen2024videocrafter2overcomingdatalimitations}, CogVideoX~\cite{yang2025cogvideoxtexttovideodiffusionmodels}, Wan2.1~\cite{wan2025wanopenadvancedlargescale}, Video-MSG~\cite{li2025trainingfreeguidancetexttovideogeneration}, and PhyT2V~\cite{xue2025phyt2vllmguidediterativeselfrefinement}. As shown in Figure~\ref{fig:human_eval}, each video was evaluated by human judges who answered two specific questions designed to assess the video’s content. The annotators followed a standardized set of instructions, shown in Figure~\ref{fig:rules}, which ensured consistency and fairness across all assessments. The evaluation focused on how well the video adhered to the given prompt and whether it accurately conveyed the intended teaching point. These human judgments provide a benchmark for comparing automatic evaluation metrics against human perception.

\begin{table}
    \centering
    \begin{tabular}{c c c c}
         Model & Duration (s)& FPS & Resolution \\
         \toprule
         VideoCrafter2~\cite{chen2024videocrafter2overcomingdatalimitations} & 5 & 8 & 512 x 320\\
         CogVideoX-5b~\cite{yang2025cogvideoxtexttovideodiffusionmodels} & 6 & 15 & 640 x 320\\
         Wan2.1~\cite{wan2025wanopenadvancedlargescale} & 6 & 15 & 832 x 480\\
         Video-MSG~\cite{li2025trainingfreeguidancetexttovideogeneration} & 6 & 28 & 720 x 480\\
         PhyT2V~\cite{xue2025phyt2vllmguidediterativeselfrefinement} & 6 & 8 & 720 x 480\\
        \bottomrule        
    \end{tabular}
    \caption{Details of duration, FPS, and resolution for each model are presented in the table.}
    \label{tab:model-details}
\end{table}

\section{Analysis of Score Mismatches Between PhyEduVideo and Human Evaluators}

\begin{figure*}[h]
    \centering
    \includegraphics[width=\textwidth,height=\textheight,keepaspectratio]{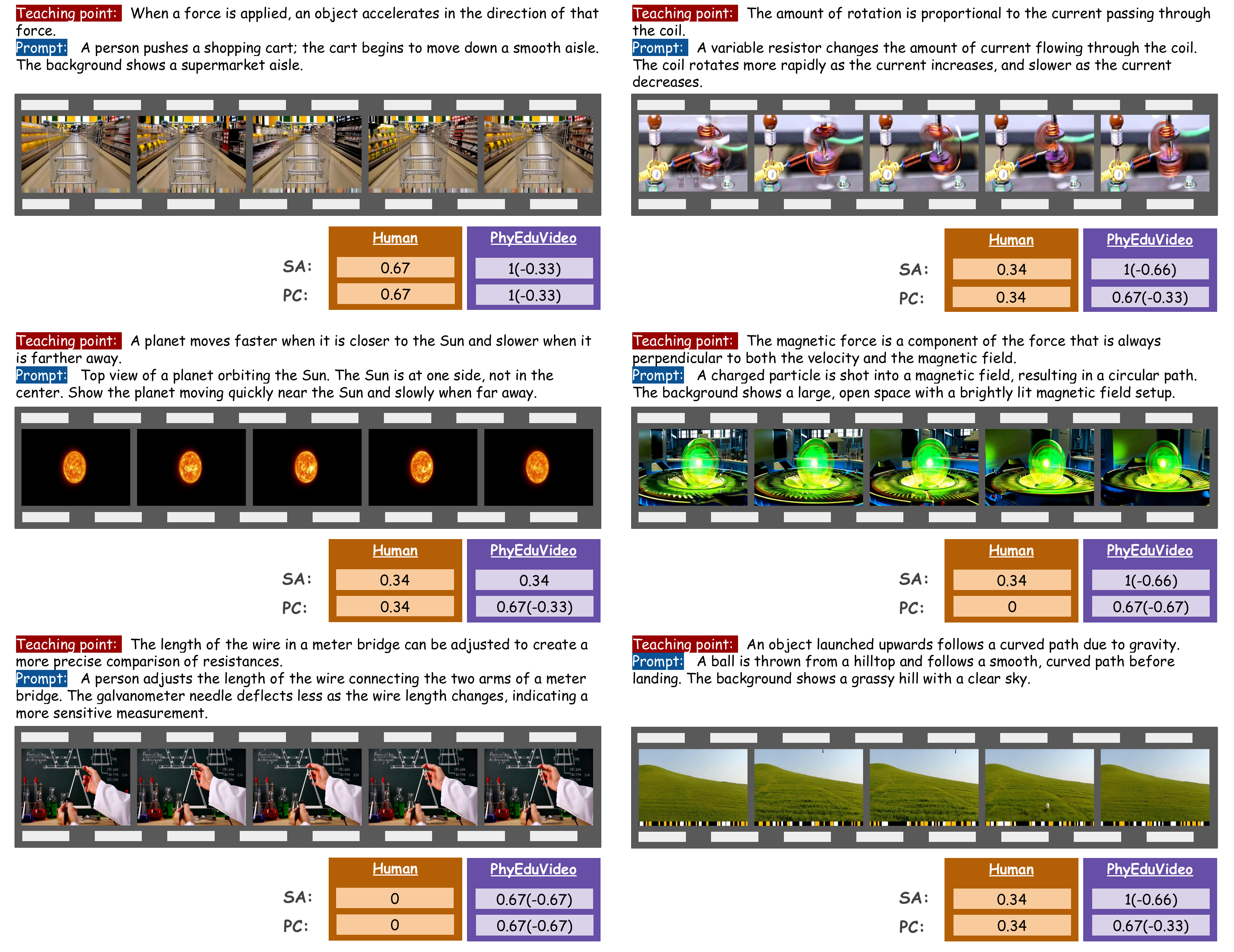}
    \caption{Comparison of SA (Semantic Adherence) and PC (Physics Commonsense) scores assigned by the Automatic Evaluator (PhyEduVideo) and humans.}
    \label{fig:failure_cases}
\end{figure*}

We analyzed cases where PhyEduVideo’s scores for Semantic Adherence (SA) and Physics Commonsense (PC) did not align with human judgments, focusing on understanding the causes of mismatches (Figure~\ref{fig:failure_cases}) . Overall, the model performs well in straightforward scenarios, such as applying force to a shopping cart, where both human and model scores perfectly match. However, in more complex cases, PhyEduVideo tends to overestimate correctness, reflecting a limitation in capturing nuanced physics reasoning or semantic context. For example, in the rotating coil scenario, humans assigned low scores (SA = 0.34, PC = 0.34) due to partial recognition of the relation between current and rotation, while PhyEduVideo overestimated both (SA = 1.0, PC = 0.67). Similarly, in planetary orbit and charged particle in a magnetic field cases, the model assigned higher scores than humans, likely because it detected general motion or field presence but failed to capture detailed physics principles, such as orbital speed variation or circular trajectories. In meter bridge wire adjustment and projectile motion on a hill, PhyEduVideo again overestimated both SA and PC, misinterpreting visual cues as correct semantic and physics adherence, whereas humans recognized subtle discrepancies in the purpose or motion. In summary, PhyEduVideo generally aligns well with human judgments for clear and straightforward scenarios. In more complex situations requiring fine-grained reasoning, it sometimes assigns higher semantic and physics scores than humans, likely due to subtle physics nuances, partial contextual cues, or reliance on visual detection of motion and objects.

\begin{figure*}[h]
    \centering
    \includegraphics[width=\textwidth,height=\textheight,keepaspectratio]{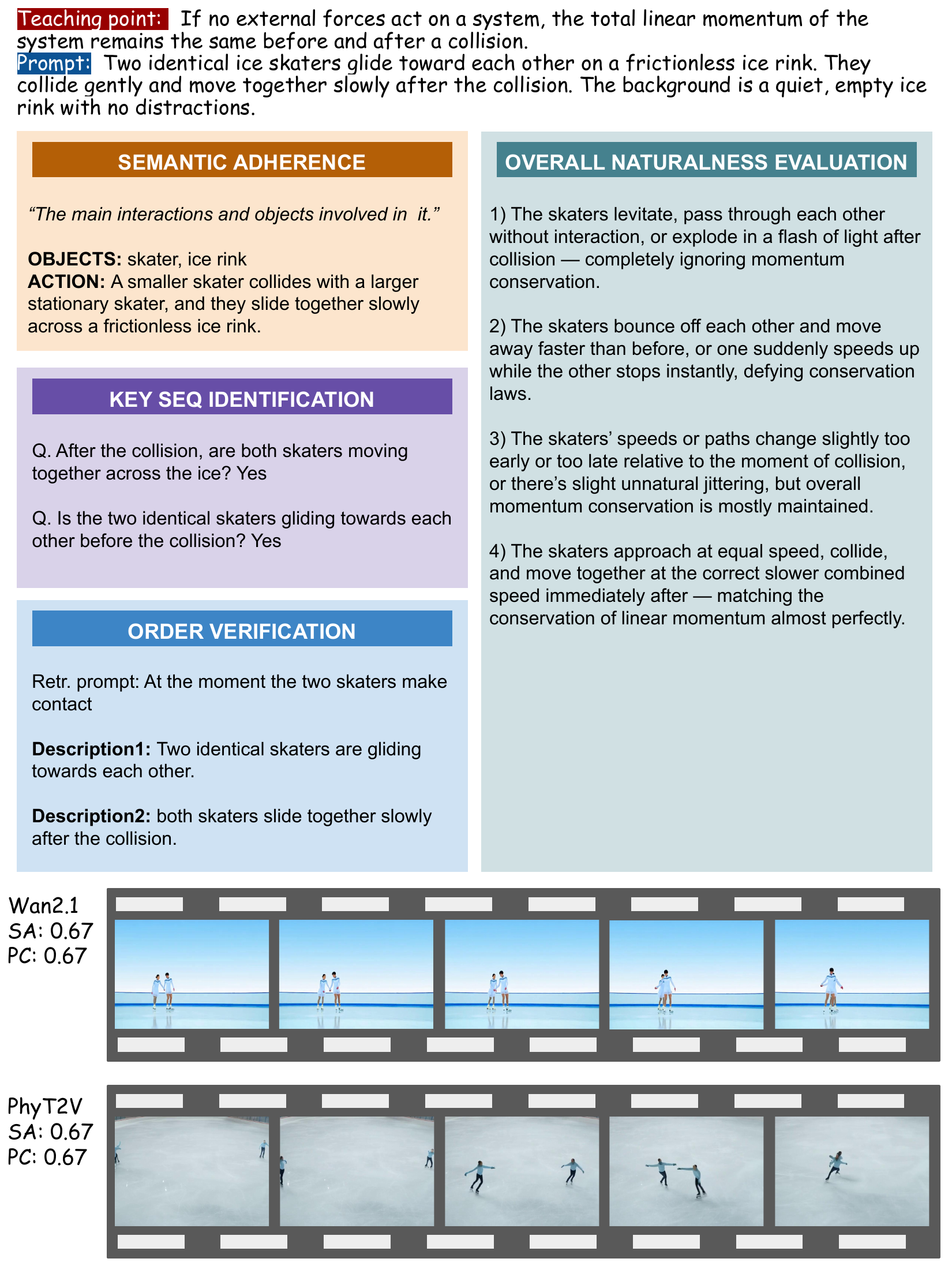}
    \caption{Domain: Mechanics. Questions used for evaluation along with outputs from Wan2.1 and PhyT2V.}
\end{figure*}

\begin{figure*}[h]
    \centering
    \includegraphics[width=\textwidth,height=\textheight,keepaspectratio]{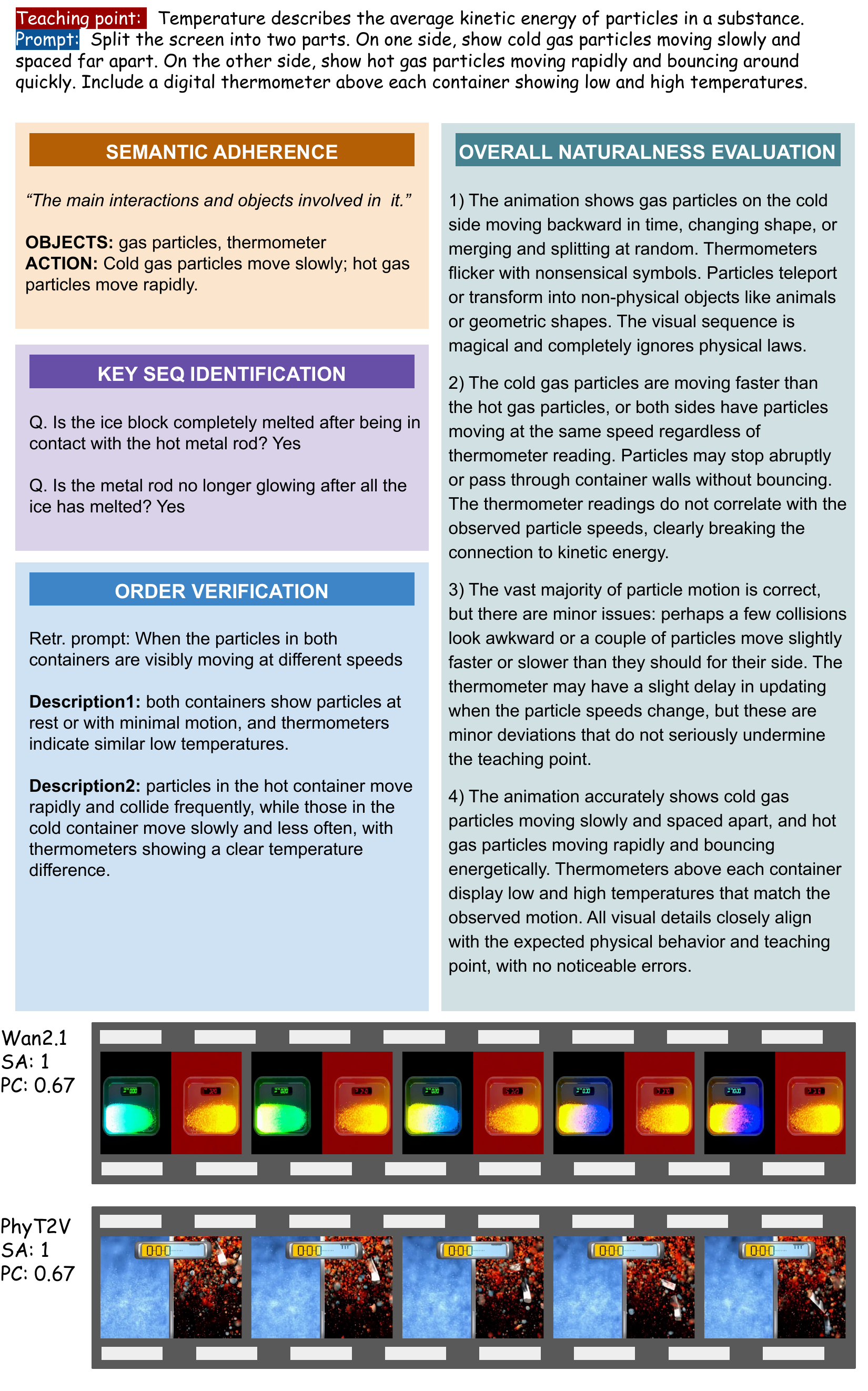}
    \caption{Domain: Thermodynamics. Questions used for evaluation along with outputs from Wan2.1 and PhyT2V.}
\end{figure*}

\begin{figure*}[h]
    \centering
    \includegraphics[width=\textwidth,height=\textheight,keepaspectratio]{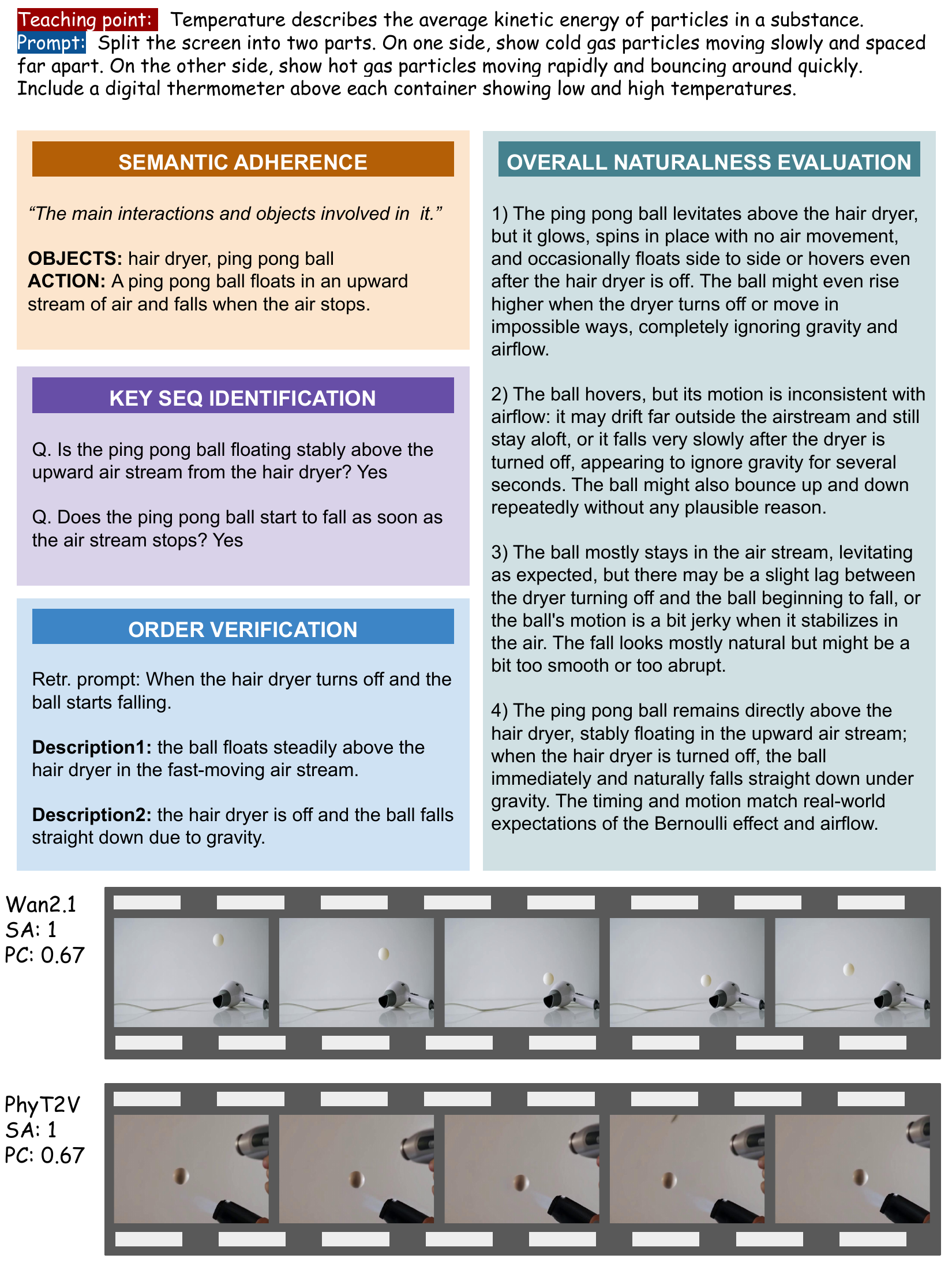}
    \caption{Domain: Fluids. Questions used for evaluation along with outputs from Wan2.1 and PhyT2V.}
\end{figure*}

\begin{figure*}[h]
    \centering
    \includegraphics[width=\textwidth,height=\textheight,keepaspectratio]{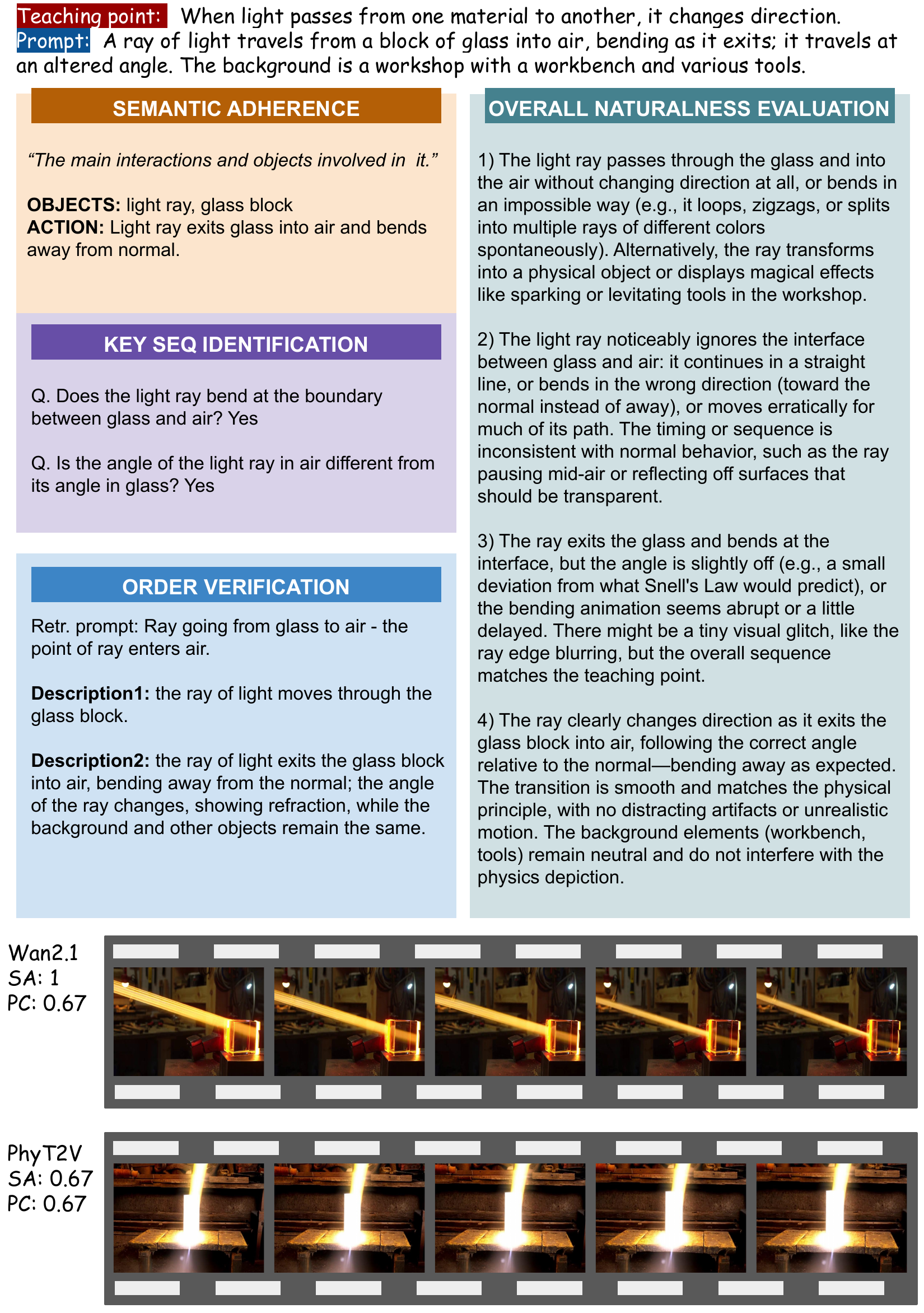}
    \caption{Domain: Optics. Questions used for evaluation along with outputs from Wan2.1 and PhyT2V.}
\end{figure*}

\begin{figure*}[h]
    \centering
    \includegraphics[width=\textwidth,height=\textheight,keepaspectratio]{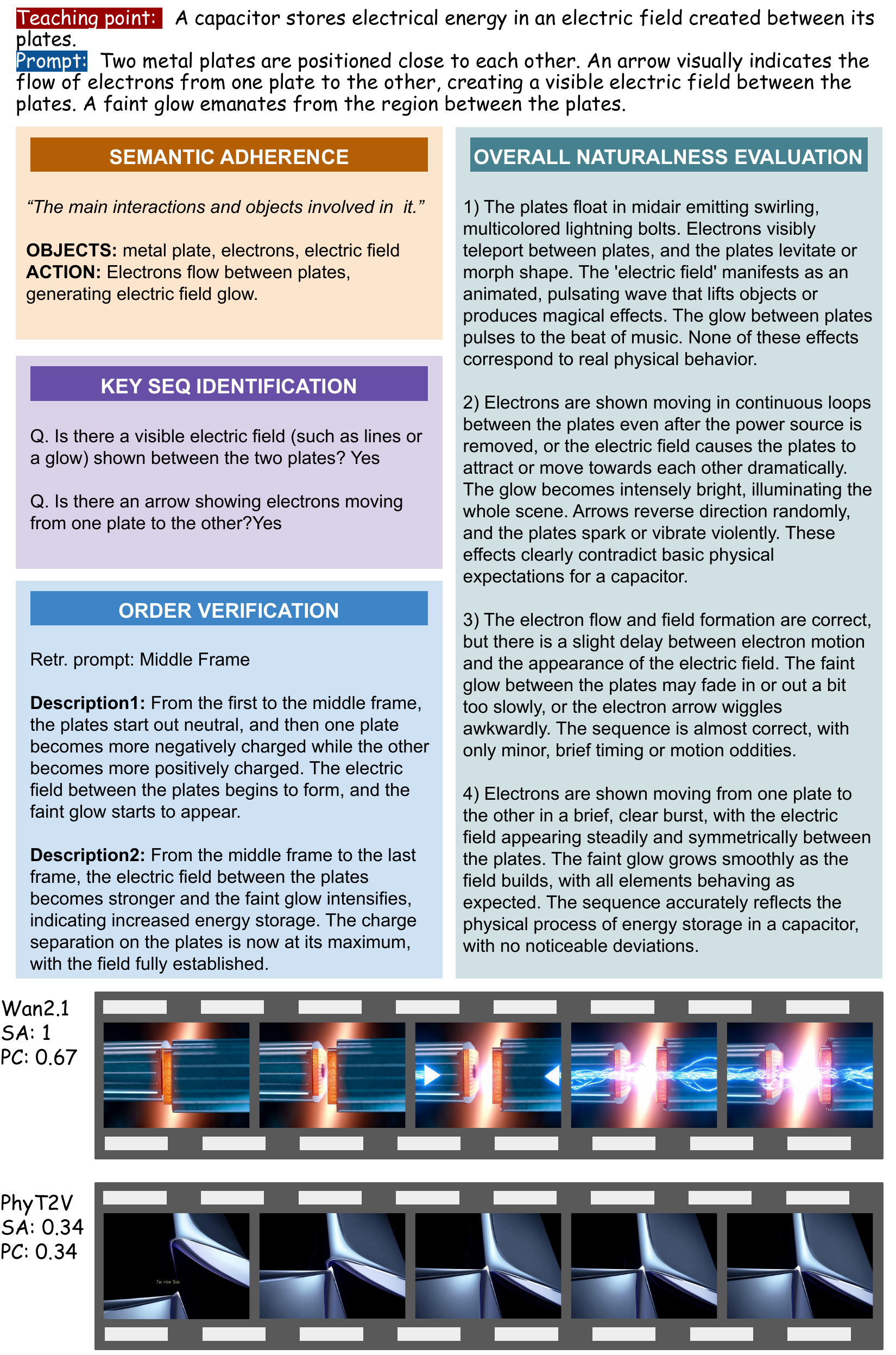}
    \caption{Domain: Electromagnetism. Questions used for evaluation along with outputs from Wan2.1 and PhyT2V.}
\end{figure*}

\begin{figure*}[h]
    \centering
    \includegraphics[width=\textwidth,height=\textheight,keepaspectratio]{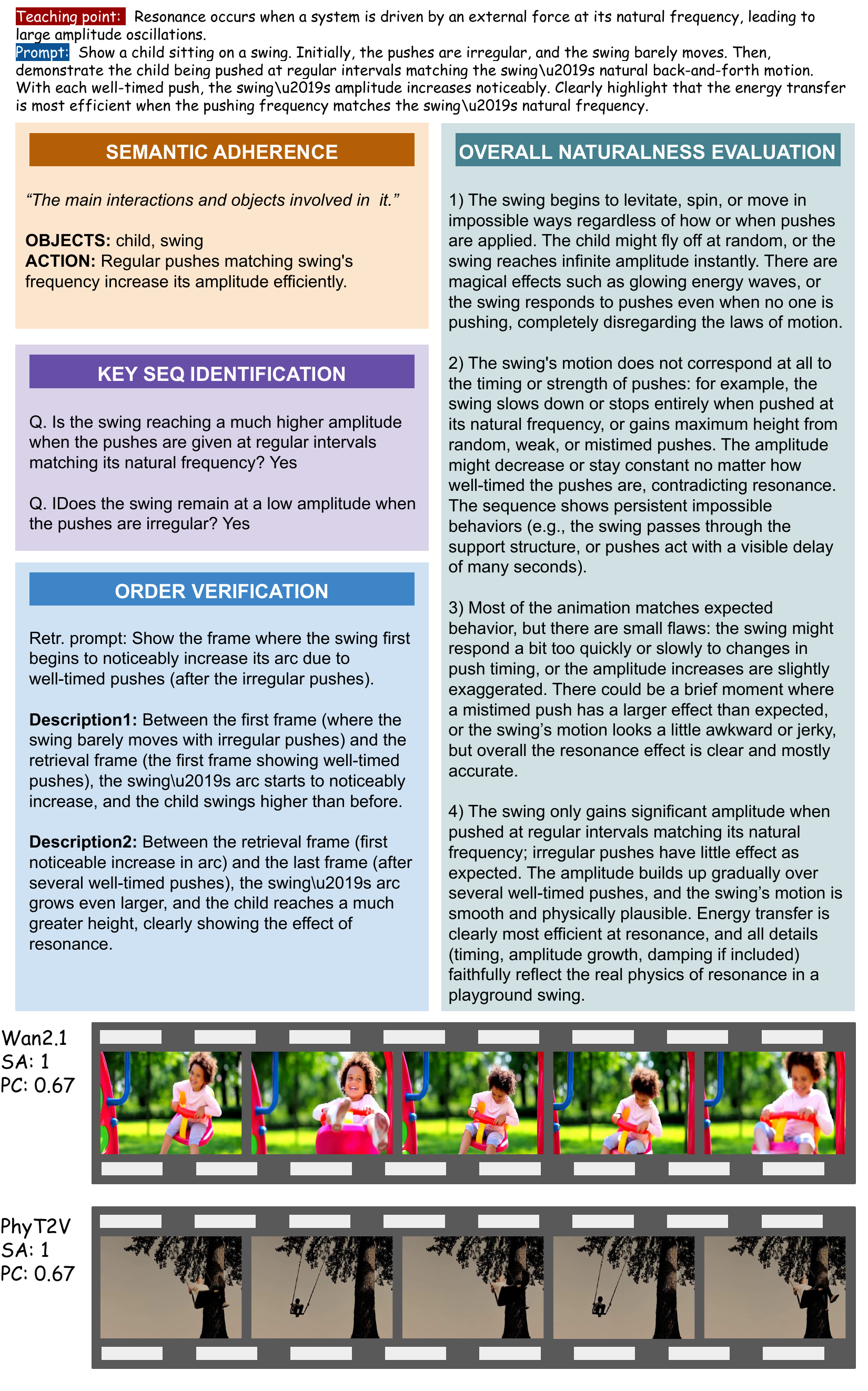}
    \caption{Domain: Waves \& Oscillations. Questions used for evaluation along with outputs from Wan2.1 and PhyT2V.}
\end{figure*}

\end{document}